\documentclass[letterpaper]{article} 
\usepackage{aaai2027}  
\usepackage[hyphens]{url}  
\usepackage{graphicx} 
\usepackage{algorithm}
\usepackage{multirow}
\usepackage{cuted}

\usepackage{algorithmic}
\usepackage{natbib}  
\usepackage{caption} 
\usepackage{pifont}
\newcommand{\cmark}{\ding{51}}%
\newcommand{\xmark}{\ding{55}}%
\usepackage{amsmath}
\usepackage{amsfonts}
\usepackage{tikz}
\usepackage{bm}
\usepackage{xcolor}

\usepackage{amssymb}
\usetikzlibrary{
    positioning,
    arrows.meta,
    fit,
    backgrounds,
    calc,
    shapes.geometric
}

\definecolor{cTeacher}{HTML}{1F77B4} 
\definecolor{cStudent}{HTML}{2CA02C} 
\definecolor{cData}{HTML}{FF7F0E}    
\definecolor{cAux}{HTML}{9467BD}     

\newcommand{\vv}{\mathbf{v}}
\newcommand{\vx}{\mathbf{x}}      
\newcommand{\vu}{\mathbf{u}}      
\newcommand{\vz}{\mathbf{z}}      
\newcommand{\vm}{\mathbf{m}}      
\newcommand{\vg}{\mathbf{g}}      
\newcommand{\ffn}{T}              
\newcommand{\ggn}{S}              
\newcommand{\pt}{p^{\ffn}}        
\newcommand{\temp}{\mathcal{T}}   
\DeclareFontFamily{U}{stix2bb}{}
\DeclareFontShape{U}{stix2bb}{m}{n} {<-> stix2-mathbb}{}

\newcommand{\etal}{\textit{et al}.}
\usepackage{booktabs}

\title{ZOPGA-KD: Zeroth Order PGA Guided Data Free Knowledge Distillation}
\author{
    Tuhin Subhra Mukherjee, Tiya Rose Pulikunnel, Vaidhav Adit, Devanand Thanissery, S.H. Shabbeer Basha
}
\affiliations{
    School of Engineering \& Technology, Vidyashilp University, Bengaluru, India.\\

}

\begin{document}

\maketitle

\begin{abstract}
Knowledge distillation (KD) enables the transfer of learned representations from a large teacher network to a small student network. In the foundational KD, the student network uses the original training data of the teacher model to accomplish this task. However, due to privacy concerns, the original training data is often not available. To overcome this reliance on original training data, there have been several data-free knowledge distillation methods developed. Some of these methods operate in the ``white-box'' setting, and a few methods generate synthetic data by training a generator network. The methods that operate in white-box settings use the parameters of the teacher model to obtain a synthetic dataset. To avoid relying on teacher parameters, some methods use computationally expensive generative approaches to obtain such a dataset. There also exist certain black-box methods that generate the synthetic data without accessing teacher parameters. This method also is computationally intensive. We propose a data-free knowledge distillation method that utilizes just the teacher's output logits and does not require access to its architecture, parameters, or gradients. We generate synthetic data via Projected Gradient Ascent (PGA) starting from smooth random noise signals. Required gradients are obtained through zeroth-order gradient estimation rather than backpropagating through the teacher's parameters. The student is then trained on these synthetically generated samples. We evaluate the proposed method on several benchmark datasets and show that the distillation performance is comparable with approaches that have access to the teacher model parameters.  
\end{abstract}



\section{Introduction}
Knowledge distillation (KD) is a framework in which the knowledge learned by a large complex network is transferred to a smaller network. The large network is referred to as the \textit{``Teacher''}, while the smaller network is referred to as the \textit{``Student''}. The teacher model delivers strong performance in its intended domain. However, its size and complexity result in a large number of parameters, making it memory-intensive and computationally slow. This makes its deployment challenging in a practical scenario. The student model, on the other hand, is much smaller and faster, making it easier to deploy.

\begin{figure*}[t]
\centering
\begin{tikzpicture}[
    font=\small,
    node distance=6mm,
    box/.style={rounded corners=5pt,draw=black,line width=0.6pt,align=center,
                text=black,minimum height=9mm,inner sep=4pt,fill=white},
    tbox/.style={box,draw=cTeacher,fill=cTeacher!8},
    sbox/.style={box,draw=cStudent,fill=cStudent!8},
    dbox/.style={box,draw=cData,fill=cData!10},
    abox/.style={box,draw=cAux,fill=cAux!8},
    kdbox/.style={box,draw=cAux!80!black,fill=cAux!10,text=black,
                  minimum height=14mm},
    stage/.style={rounded corners=5pt,draw=black,dashed,line width=0.6pt,
                  inner sep=8pt},
    fl/.style={-{Stealth[length=2.5mm]},thick,black},
]
\node[dbox] (noise) {Smooth Gradient Noise\\initialization $\vx$};
\node[abox,right=of noise] (zo) {ZO-PGA};
\node[tbox,above=8mm of zo] (t1) {Teacher $T$\\(black box)};
\node[dbox,right=14mm of zo] (synth) {Synthetic set\\$\mathcal{D}$\\(images,\,soft labels)};
\draw[fl] (noise) -- (zo);

\draw[fl] ([xshift=-3mm]zo.north) -- node[left,font=\small,text=black]{query}
          ([xshift=-3mm]t1.south);
\draw[fl] ([xshift=3mm]t1.south) -- node[right,font=\small,text=black]{logits}
          ([xshift=3mm]zo.north);
\draw[fl] (zo) -- (synth);
\node[kdbox,right=16mm of synth] (kd) {Knowledge\\Distillation};
\node[sbox,right=of kd] (out) {Student $S$};
\draw[fl] (synth) -- (kd);
\draw[fl] (kd) -- (out);
\begin{scope}[on background layer]
  \node[stage,fit=(noise)(zo)(t1)(synth),
        label={[black,font=\small\bfseries]above: Stage 1: ZO-PGA synthesis}] {};
  \node[stage,fit=(kd)(out),
        label={[black,font=\small\bfseries]above:  Stage 2: Student trained using DFKD}] {};
\end{scope}
\end{tikzpicture}
\caption{End-to-end data-free knowledge distillation pipeline. In Stage~1, the teacher $\ffn$ is
accessed only through forward queries (finite differences) to
synthesize a soft-labeled transfer set $\mathcal{D}$ from noise; in Stage~2 the
student $\ggn$ is trained on this synthetic set using knowledge distillation with MSE loss function.}\label{fig:overview}
\end{figure*}
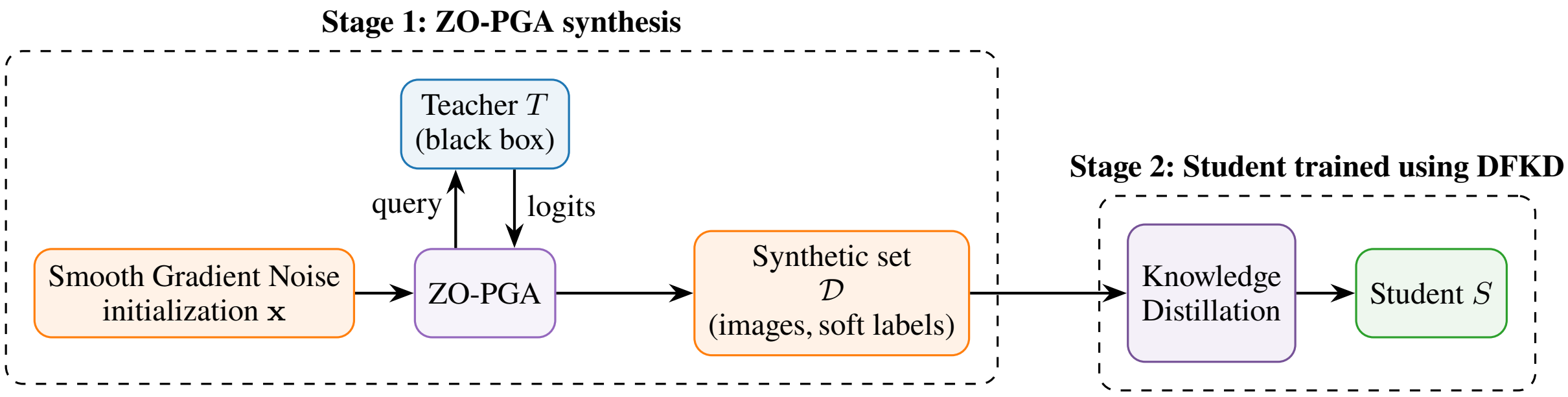

The original work by \citet{hinton2015distilling} recommends using the original data (used to train the teacher) while distilling knowledge to the student. But due to privacy concerns and/or data regulations, in a few settings, the original data may not be available for public use. This gives rise to the paradigm of data-free knowledge distillation (DFKD), which does not use the original training data for KD\@. Instead, the KD is performed on a synthetic dataset. However, an arbitrary dataset cannot be used; the distribution of the synthetic dataset should closely approximate the distribution of the original data. One of the seminal works~\cite{yin2020dreaming} in this direction makes use of teacher network batch-norm statistics to synthesize the dataset.\ \citet{nayak2019zero} use the weights of the final layer, and \citet{wang2021data} utilize the weights of all the layers of the teacher to construct the synthetic dataset. Since these methods need access to the teacher's parameters, they are generally referred to as ``white-box'' methods. In certain scenarios, we neither have access to the teacher's parameters nor the architecture. In such cases, the only way to interact with the model is to query the teacher with some inputs and record the outputs. To facilitate DFKD in such a ``black-box'' setting, several works~\cite{micaelli2019zero,truong2021data,hong2025data,chen2019data,kariyappa2021maze,fang2022up,zhang2022ideal,fang2019data,braun2024deep} employ generative frameworks to obtain the synthetic dataset. However, these approaches introduce an additional overhead since a generator network has to be trained explicitly.


\citet{wang2021zero} proposed a black-box method that generates synthetic data by pushing random noise signals away from the teacher's decision boundaries. The required soft labels are constructed using the concept of robustness, which they obtain by calculating the distance of the samples from decision boundaries. This method, however, is also computationally expensive since it requires 4000 million queries~\cite{yuan2024data}.

In this work, we explore an efficient black-box, generator-free method called ZOPGA-KD, which uses only the output logits of the teacher; no other information about the teacher, such as parameters or architecture, is used. To be concrete, we generate the synthetic dataset using Projected Gradient Ascent (PGA) in the input space, starting from the random noise signals. We avoid backpropagation through the teacher by zeroth-order gradient approximation~\cite{nesterov2017random}. The generated samples, aided by data augmentation, are then used to train the student using the Mean Squared Error (MSE) instead of Kullback–Leibler (KL) divergence as suggested in \citet{kim2021comparing}. We demonstrate the efficacy of our method by applying it to several teacher and student architectures across various benchmark datasets. We compare ZOPGA-KD results with state-of-the-art, generator free, white-box and black-box methods. Note that Zeroth order optimization is previously used in generator based methods of DFKD (\citet{kariyappa2021maze},\citet{truong2021data}) to update the parameters of the generator. We highlight our contributions as follows:
\begin{itemize}
    \item We propose a generator-free, black-box, DFKD method (``ZOPGA-KD'') which performs KD by generating synthetic datasets through PGA on the input space using zeroth-order gradient approximation. This method avoids both the need for the original data and access to the parameters/architecture of the teacher network. \citet{wang2021zero} also usues zeroth order optimization on the input space, operating in the hard-label setting. They start from random noise signals and iteratively push them away from the estimated decision boundaries. The soft-label for each of the samples is approximated by measuring its distance from the relevant decision boundaries, which is an iterative procedure. This method incurs a high computational cost, requires a large number of queries, and is challenging to extend to more complex datasets such as CIFAR-100 and TinyImageNet \cite{yuan2024data}.
    

    \item We show that the simple mean squared error (MSE) loss effectively performs the knowledge distillation. 
    \item By applying the proposed method on various standard architectures and datasets, we demonstrate the feasibility of our approach by comparing with most of the existing works in the literature. 
\end{itemize}

\section{Related Works}

As mentioned in the previous section, KD seeks to train a small student network to mimic the behavior of a large complex teacher network. The method described in the foundational work of \citet{hinton2015distilling} is as follows. Let $f$ be a classifier designed to classify inputs into $K$ classes. It takes an input sample $x$ and produces a $K$-dimensional representation (logits), $\hat{f}(x)=z\in \mathbb{R}^K$, which, after applying the softmax function, gives the softmax vector $F(x)$ with components $F_i(x)=e^{z_i}/\sum_{i=1}^K e^{z_i}$. Let us denote the teacher and the student networks by $T$ and $S$, respectively. Given an input $x$, the teacher's output softmax vector $T(x)$ not only tells us about the class that $x$ belongs to but also gives information about the relative similarity between the classes, the so-called ``dark knowledge''. To exploit this dark knowledge, the student is trained to approximate the distribution of the softmax vectors of the teacher. This is accomplished by minimizing the KL divergence:
\begin{align}\label{KLD.Eq}
    \mathcal{L}_{KL}(T,S)=\frac{1}{N}\sum_{j=1}^N \sum_{i=1}^K T_i(x_j) \ln \left(\frac{T_i(x_j)}{S_i(x_j)}\right)
\end{align}
where $N$ is the number of training samples and $K$ is the number of classes. However, for most inputs, the softmax vector has one dominant component (close to $1$) while others are very small. In that case, the loss in Eq.~(\ref{KLD.Eq}) reduces to the usual cross-entropy, in which the teacher provides the hard labels. To cure this, the softmax probabilities are softened by scaling the raw logits with temperature $\temp$ as $z\to z/\temp$. 

In \citet{hinton2015distilling}, original training data is utilized for the distillation process. Therefore, they have also added the cross-entropy term to the loss function, giving the total loss for the student network
\begin{align}\label{KL-CE.Eq}
    \mathcal{L}=\mathcal{L}_{KL}(T,S)- \frac{1}{N}\sum_{j=1}^N\sum_{i=1}^K y_i(x_j) \ln S_i(x_j)
\end{align}

where $y(x_j)$ is the one-hot encoding vector of the true label for the sample $x_j$. for the input sample $x_j$. It turns out that training the student with the loss in Eq.~(\ref{KL-CE.Eq}) is much more efficient than training only with the hard labels (through the cross-entropy loss). 

Due to privacy or security regulations, the original training data may be unavailable. This calls for data-free knowledge distillation (DFKD). The existing literature can be broadly classified into two categories according to the teacher information used during distillation: (a) white-box methods, which require the teacher's parameters, gradients, or architectural details, and (b) black-box methods, which access only the teacher's output logits.

\subsection{White-box methods for DFKD}
The work of \citet{yin2020dreaming} proposes ``deepInversion'', which synthesizes images by starting with random images and nudging them towards realistic images by minimizing the cost function 
\begin{align*}
    \mathcal{L}=\mathcal{L}_{CE}(x_i,y_i)+\mathcal{L}_{Reg}
\end{align*}
where $x_i$ is the input image and $y_i$ is the one-hot encoding vector for the $i^{th}$ training sample. Here, $\mathcal{L}_{Reg}$ is the regularizer, which is responsible for the realistic nature of the images. It penalizes the images for which the layer-wise feature statistics differ from those of the original data. The statistics of the original data are estimated by the batch-norm statistics of the teacher stored in its batch-norm layers. In the same work, they have also developed a variation of deepInversion (known as ``adaptive deepInversion'') that uses a generator to synthesize images that cause the teacher to disagree with the student. 

In \citet{chen2019data}, the authors use a generator to produce images whose distribution is closer to the original distribution. They achieve this by defining a generator loss that encourages (a) one-hot-like teacher softmax vectors, (b) high $\ell_1$ norms for feature vectors in the layer before the fully connected layer, and (c) a balanced dataset. The student is trained by KL divergence on the synthetic data.

\citet{nayak2019zero} (``\textbf{ZSKD}'') obtained the synthetic data by using the weights of the final fully connected layer of the teacher. They modeled the distribution of the softmax vectors for class $i$ as a Dirichlet distribution, $\operatorname{Dir}(K, \alpha_i)$, where $\alpha_i$ is a concentration-parameter vector that determines the similarity of class $i$ to the other classes. In ZSKD, the vectors $\alpha_i$ are determined from vectors $c_i$ whose $j$-th component is
\begin{align*}
    c_{ij}=\frac{W_i^{\mathsf{T}}W_j}{\lVert W_i\rVert_2\lVert W_j\rVert_2}
\end{align*}
Here, $W_i$ is the vector of weights connecting the penultimate layer to the $i$-th neuron of the final layer. The intuition is that $c_{ij}$ captures the relative similarity of classes $i$ and $j$ as perceived by the teacher. To generate the data ($\bar{x}$), they sampled softmax vectors $y_i$ from $\operatorname{Dir}(K, \alpha_i)$ and solved the following equation:
\begin{align*}
    \bar{x}=\arg\!\min_x \mathcal{L}_{CE}(y_i,T(x))
\end{align*}
The student is trained on the generated synthetic dataset using KL divergence between the student and the teacher.

\citet{wang2021data} takes a similar approach to that of ZSKD but uses a multivariate Gaussian distribution to model the softmax distribution. They also use the weight vectors from all layers, rather than only the final layer, for better estimation. Additionally, they try to maximize the $\ell_1$ norm of the output feature maps for all layers by incorporating a regularizer to the loss function, similar to \citet{chen2019data}.

\subsection{Black-box methods for DFKD}
In this section, we will discuss some of the DFKD methods in the black-box setting.

In the work of \citet{kariyappa2021maze}, DFKD is performed by training a generator together with the student. Starting from random noise vectors ($z$), the generator $G$ produces images that are given as input to both the teacher and the student networks. During training, the student parameters are updated to produce outputs that match the teacher, while the generator is trained to produce images for which the student disagrees with the teacher. To update the generator parameters $\theta_G$, the gradient of the generator loss $\mathcal{L}_G$ factors as
\begin{align*}
   \frac{\partial \mathcal{L}_G}{\partial \theta_G}=\frac{\partial \mathcal{L}_G}{\partial x}\frac{\partial x}{\partial \theta_G}
\end{align*}
The gradient $\frac{\partial \mathcal{L}_G}{\partial x}$ is calculated via zeroth-order gradient approximation, which does not require the teacher gradients. \citet{truong2021data} uses the same method as Kariyappa \etal, except that mean absolute error (MAE) is used as the loss function for student training.

The work of \citet{zhang2022ideal} also uses a generator but in a different way than the works mentioned previously. They iteratively train the generator and the student in two stages. The first stage generates synthetic data, and in the second stage, the student network is trained using the synthetic data generated. During the data generation stage, the generator is trained to produce synthetic images (from noise signals) of desired classes by querying the \textit{fixed} student model. In the second stage, the student is trained on the synthetic data by minimizing $\mathcal{L}_{KL}(T,S)$.

\citet{wang2021zero} presented another method (\textbf{ZSDB3KD}) that uses only the hard labels of the teacher. They started with random noise signals and gradually pushed them away from the targeted decision boundary using gradient ascent which they calculated through zeroth-order gradient estimation. They assigned soft labels to the synthetic images by calculating minimal boundary distances from the decision boundaries. Later, the generated synthetic images to train the student network.


\section{Methodology}\label{sec:method}

\begin{algorithm}[t]
\caption{ZO-PGA Synthesis of One Sample (black box)}\label{alg:zopga}
\begin{algorithmic}[1]
\REQUIRE{} Black-box teacher $\ffn$ (soft labels via queries only); target class $c$;
  noise distribution $\mathcal{P}$ on ${[0,1]}^{C \times H \times W}$;
  confidence threshold $\tau$; iteration cap $t_{\max}$; smoothing parameter
  $\sigma$; directions per step $q$; perturbation resolution
  $r \le \min(H, W)$; step size $\alpha$; momentum $\beta$
\ENSURE{} Sample $(\vx, \vz)$ with $\vz = \ffn(\vx)$
\STATE{} $J_c(\vx) \leftarrow
  \ln \big(\ffn_c(\vx)\big)$ \COMMENT{Objective; $\nabla_{\vx} J_c$ unavailable}
\STATE{} $\vx \sim \mathcal{P}$; \; $\vm \gets \bm{0}$; \; $t \gets 0$
\WHILE{$\pt_c(\vx) < \tau$ \textbf{and} $t < t_{\max}$}
  \STATE{} $J^{0} \gets J_c(\vx)$
  \FOR{$i = 1, \dots, q$}
    \STATE{} $\vv_i \sim \mathcal{N}(\bm{0}, I)$ of shape $C \times S \times S$
    \STATE{} $\vu_i \gets \mathrm{Upsample} ~\vv_i$ to $C\times H \times W$
  \ENDFOR{}
  \STATE{} $\widehat{\vg} \gets \dfrac{1}{q} \displaystyle\sum_{i=1}^{q}
    \dfrac{J_c(\vx + \sigma \vu_i) - J^{0}}{\sigma}\, \vu_i$    
  \STATE{} $\vm \gets \beta\, \vm + (1 - \beta)\, \widehat{\vg}$ 
  \STATE{} $\vx \gets \mathrm{clip}_{[0,1]}
    \big(\vx + \alpha\, \vm / \|\vm\|_2\big)$; \; $t \gets t + 1$
\ENDWHILE{}
\RETURN{} $\vx, \ffn(\vx)$
  \COMMENT{synthetic sample, soft labels}
\end{algorithmic}
\end{algorithm}

The outline of the proposed Zeroth-Order Projected Gradient Ascent guided data-free Knowledge Distillation (ZOPGA-KD) is presented in Fig.~\ref{fig:overview}. The proposed method does not require access to teacher model weights or gradients; however, other knowledge distillation methods~\cite{nayak2019zero,wang2021data} require such access. In the proposed method, the teacher network is utilized only to obtain the logits. Similar to \citet{nayak2019zero,wang2021zero}, our method proceeds in two \textit{independent} stages: i). Data Generation, ii). Student Training.

\subsection{Data Generation}
Suppose we want to distill the knowledge of a teacher $T$ trained to classify images into $K$ classes. For a given input $x$, the output confidence vector (or softmax vector) is denoted by $T(x)$.

Our goal is to synthesize $M$ images per class. We first fix an upper bound on the confidince score $\tau$. Then, for each class $i$, we sample a batch of $B$ initial random noise images $\in {[0,1]}^{C\times H\times W}$ from a chosen distribution $\mathcal{P}$ at each iteration. Here, $C$, $H$, and $W$ denote the number of channels, height, and width of the image, respectively. We optimize these noisy images using Projected Gradient Ascent (PGA)~\cite{salvatore2026solving} as follows. Let $T_i(x)$ be the confidence score for the $i$-th class for an input $x$. In every iteration of PGA, the pixels of $x$ are updated as
\begin{align}\label{optim.Eq}
    x_{t+1}=x_{t} + \alpha \frac{\nabla_x\left(\ln T_i(x)\right)}{||\nabla_x\left(\ln T_i(x)\right)||_2}
\end{align}
where $\alpha$ is the step size. After one or a few iterations, some pixels might fall outside the range $[0,1]$. We therefore project them back to $[0,1]$ by clipping, as shown in Algorithm~\ref{alg:zopga}. We iterate this process until $T_i(x)$ reaches the desired value $\tau$ or a maximum of $t_{\max}$ iterations has been performed, whichever happens first. If we had access to the teacher's parameters, we could calculate the gradient by backpropagating through the teacher. The proposed ZOPGA-KD performs knowledge distillation in a black-box setting. Hence, we estimate the gradient in Eq.~(\ref{optim.Eq}) via zeroth-order gradient approximation~\citep{nesterov2017random}. The proposed ZOPGA algorithm for data generation is presented in Algorithm~\ref{alg:zopga}. 

Zeroth-order gradient approximation (ZOA) estimates the gradient of a function $f$ using only function evaluations at selected inputs $x$. We sample $q$ random perturbation directions $v_i$ from a $C\times S\times S$-dimensional Gaussian distribution with zero mean and unit variance, where $S\leq H,W$.
We then bilinearly upsample them to the required $C\times H\times W$ directions $u_i$. Upsampling the low-resolution perturbations $v_i$ reduces the number of queries~\citep{tu2019autozoom}. The gradient is then estimated as in \citet{ilyas2018black}:
\begin{align*}
\hat{g}(x)=\nabla_x J(x) \approx \frac{1}{q}\sum_{i=1}^{q}\frac{J(x+\sigma u_i)-J(x)}{\sigma}u_i
\end{align*}
where $\sigma$ is the Gaussian smoothening parameter. In fact, we use an adaptive upsampling \citep{ilyas2018prior}. Initially, when the images have low confidence ($<0.5$), we consider lower resolution $C\times S_1 \times S_1$ noise which is upsampled to $C\times H \times W$. On the other hand, if the confidence score reaches $0.5$, we start with relatively higher resolution $C\times S_2 \times S_2$, is upsampled to required dimension i.e., $C\times H \times W$. In our experiments, for gray-scale images,  $S_1=4$ and $S_2=7$ are used, while for RGB images, $S_1=4$ and $S_2=8$ are used. This process will reduce the number of queries.

At step $j$, the velocity $m_j$ is defined as an exponential moving average (EMA), which smooths the noisy ZOA updates resulted using Eq.~(\ref{optim.Eq}):
 
\begin{align}
    m_{j+1}=\beta m_j +(1-\beta) \hat{g}(x_j)
\end{align}
where $\beta$ is called the momentum which is a hyperparameter.
Then the update rule is modified as
\begin{align}
    x_{j+1}= x_j +\alpha \frac{m_{j+1}}{||m_{j+1}||_2}
\end{align}

We also optimize GPU usage in several ways to speed up the data generation pipeline. Instead of optimizing all $M\times K$ images at once, we maintain a persistent pool of $B$ slots. In each iteration, every image is optimized once. Once an image surpasses the confidence threshold $\tau$ or reaches the maximum of $t_{\max}$ iterations, it retires. A new random noise image is then launched into the vacant slot for the class with the largest number of images still to be optimized.
Moreover, if optimization stalls for an image, it is removed from the optimization protocol and the slot is reinitialized with fresh noise. Specifically, if an image's confidence score remains below $0.3$ after $300$ iterations, it is abandoned and replaced by another noise image.

\begin{figure}[t]
\centering    
\resizebox{\columnwidth}{!}{%
\begin{tikzpicture}[
    node distance=0.6cm and 1.0cm,
    >=Latex,
    box/.style={
        draw=black,
        fill=white,
        rounded corners=5pt, 
        minimum height=0.95cm, 
        minimum width=4.2cm, 
        align=center, 
        font=\small,
        line width=0.6pt
    },
    databox/.style={
        box,
        draw=cData,
        fill=cData!10
    },
    auxbox/.style={
        box,
        draw=cAux,
        fill=cAux!8
    },
    smallbox/.style={
        box,
        minimum height=0.8cm,
        minimum width=2.8cm
    },
    teacherbox/.style={
        smallbox, 
        draw=cTeacher, 
        fill=cTeacher!8
    },
    studentbox/.style={
        smallbox, 
        draw=cStudent, 
        fill=cStudent!8
    },
lossbox/.style={
    smallbox,
    draw=cAux!80!black,
    fill=cAux!10,
    minimum width=3.8cm,
    text=black
},
    arrow/.style={
        ->, 
        thick, 
        black, 
        line join=round
    },
    backprop/.style={
        ->, 
        thick, 
        dashed, 
        black, 
        line join=round
    }
]

\node[databox] (input) {Synthetic images $\textbf{X}$};
\node[auxbox, below=of input] (basegeo) {Base geometric transform\\[0.1em]\footnotesize Reflect pad + random flip};
\node[auxbox, below=of basegeo] (aug) {$n_{ops}$ random ops\\[0.1em]\footnotesize Selected from pool of size $n_{tot}$};
\node[auxbox, below=of aug] (norm) {Normalize\\[0.1em]\footnotesize $(x-\mu_{\mathrm{synth}})/\sigma_{\mathrm{synth}}$};

\node[teacherbox, below=0.8cm of norm, xshift=-1.8cm] (teacher) {\textbf{Teacher $T$}\\[0.1em]\footnotesize (frozen)};
\node[studentbox, below=0.8cm of norm, xshift=1.8cm]  (student) {\textbf{Student $S$}\\[0.1em]\footnotesize (trainable)};

\node[teacherbox, below=0.6cm of teacher] (Tout) {Teacher logits\\[0.1em]\footnotesize $\hat{t}(x)$};
\node[studentbox, below=0.6cm of student] (Sout) {Student logits\\[0.1em]\footnotesize $\hat{s}(x)$};

\path (Tout) -- (Sout) coordinate[midway] (logitsmid);
\node[lossbox, below=0.8cm of logitsmid] (loss) {\textbf{MSE loss}\\[0.1em]\footnotesize $\mathcal{L}_{\mathrm{MSE}} = \frac{1}{N}\sum_x \|\hat{t}(x)-\hat{s}(x)\|_2^2$} ;

\draw[arrow] (input) -- (basegeo);
\draw[arrow] (basegeo) -- (aug);
\draw[arrow] (aug) -- (norm);

\draw[arrow] (norm.south) -- ++(0,-0.35) -| (teacher.north);
\draw[arrow] (norm.south) -- ++(0,-0.35) -| (student.north);

\draw[arrow] (teacher) -- (Tout);
\draw[arrow] (student) -- (Sout);

\draw[arrow] (Tout.south) -- (loss.north -| Tout);
\draw[arrow] (Sout.south) -- (loss.north -| Sout);

\draw[backprop] (loss.east) to[out=0, in=-25, looseness=1.4] 
    node[midway, right=2pt, font=\scriptsize\bfseries, align=left] {Backprop\\[-2pt](student only)} 
    (student.east);

\end{tikzpicture}
}
\caption{Schematic of the student training pipeline.}\label{fig:student_training}

\end{figure}
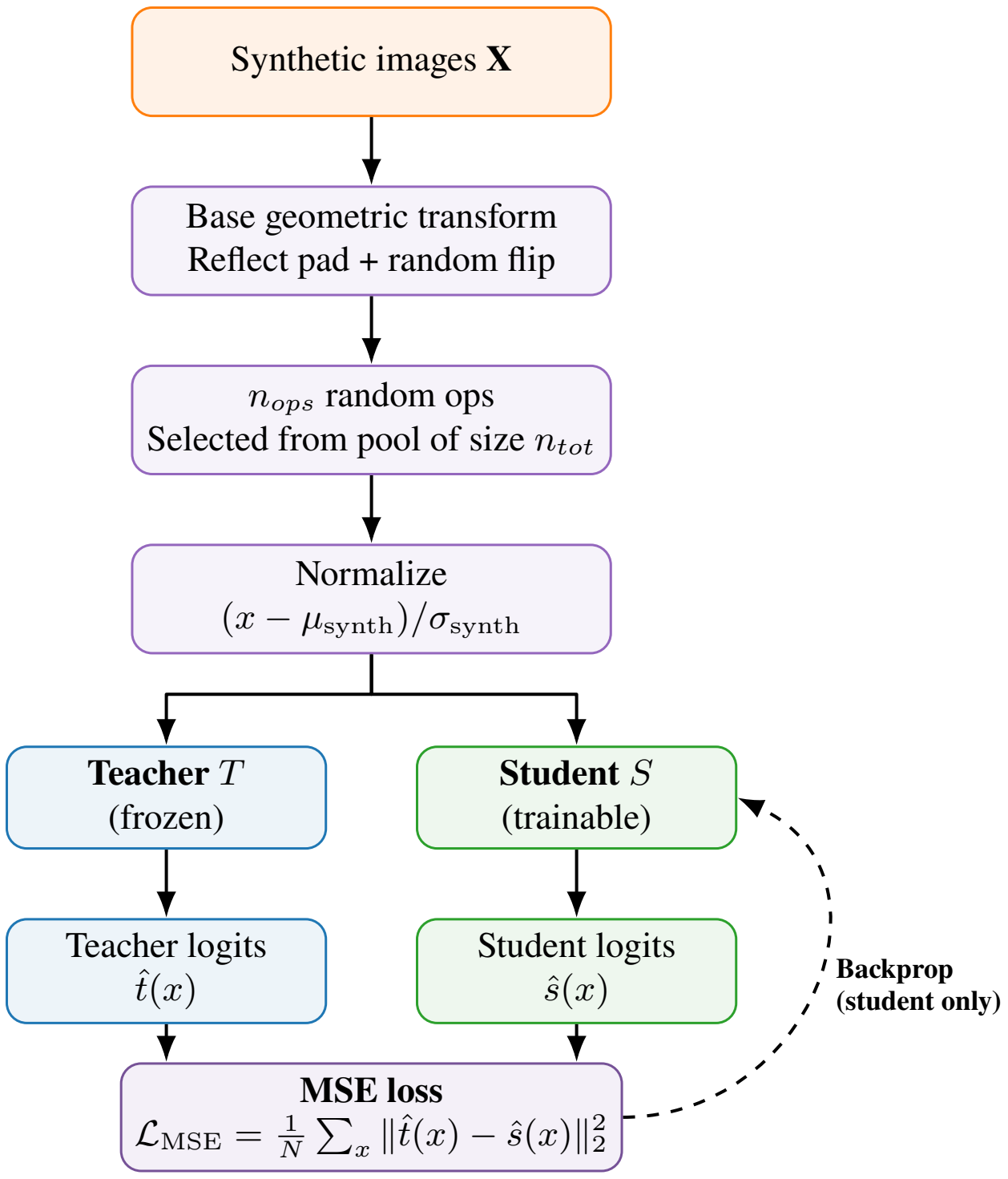

\begin{table*}[t]
\centering
\small
\setlength{\tabcolsep}{6pt}
\caption{Student test accuracy (\%) for several data-free knowledge distillation techniques and baseline values.
\emph{Access}: WB indicates white-box in which teacher parameters are accessed, whereas BB indicates black-box which does not has access to teacher parameters.
\emph{Data}: \cmark{} indicates that a method uses real training data and is therefore not data-free; \xmark{} indicates a fully data-free method. Query estimates for ZOPGA-KD method is also given.
Generation queries are consumed while synthesizing $N_{\mathrm{synth}}$ samples.
Augmentation queries are the second-stage cost during student training and equal $200\times N_{\mathrm{synth}}$, since training runs for 200 epochs. The query cost for ZSDB3KD, reported here, is a lower bound. All of the methods use Adam optimizer. ZSKD and \citet{wang2021data} use the same temperature ($\temp=20$) as us. On the other hand, ZSDB3KD uses $\temp=0.3$.}\label{tab:main-results}
\begin{tabular}{lccrrrr}
\toprule
& & & MNIST & F-MNIST & \multicolumn{2}{c}{CIFAR-10} \\
\cmidrule(lr){4-4}\cmidrule(lr){5-5}\cmidrule(lr){6-7}
Method & Access & Data & LeNet-5 & LeNet-5 & AlexNet & RN-34/18 \\
\midrule

Teacher (CE)                    & \textemdash{} & \cmark{} & 99.00 & 90.32 & 84.95 & 95.31 \\
Student (CE)                    & \textemdash{} & \cmark{} & 99.04 & 90.44 & 84.76 & 94.23 \\
Classical KD                    & \textemdash{} & \cmark{} & 99.00 & 88.47 & 82.85 & 94.28 \\
DB3KD-MBD~\cite{wang2021zero}   & BB & \cmark{} & 99.52 & 91.45 & 78.30 & 77.31 \\

ZSKD~\cite{nayak2019zero}       & WB & \xmark{} & 98.77 & 79.62 & 69.56 & \textemdash{} \\
Wang et al.~\cite{wang2021data} & WB & \xmark{} & 99.08 & \textemdash{} & 73.91 & 93.31 \\
ZSDB3KD~\cite{wang2021zero}     & BB & \xmark{} & 96.54 & 72.31 & 59.46 & \textemdash{} \\
\textbf{ZOPGA-KD (ours)}        & BB & \xmark{} & \textbf{97.40} & \textbf{80.06} & \textbf{78.41} & \textbf{87.97} \\
\midrule
\multicolumn{7}{l}{\emph{ZOPGA-KD teacher-query cost}} \\
\quad Synthetic samples $N_{\mathrm{synth}}$ & & & 24K & 48K & 40K & 40K \\
\quad Generation queries       & & & 342.18M   & 754.74M & 192.44M & 36.14M \\
\quad Augmentation queries     & & & 4.80M  & 9.60M   & 8.00M   & 8.00M  \\
\midrule
\quad \emph{ZSDB3KD teacher-query cost}&~ & ~&2000M & 2000M & 4400M &\textemdash{} \\
\bottomrule
\end{tabular}
\end{table*}

\subsection{Student Training}
Having generated the synthetic data, we now proceed to train the student. It is standard in literature to train the student using the KL-divergence given in Eq.~(\ref{KLD.Eq}) as it was originally proposed. However as demonstrated in \citet{kim2021comparing}, simple Mean Squared Error (MSE) outperforms the KL divergence in certain scenarios. Inspired by this result we use the MSE loss
\begin{align}
    \mathcal{L}(T,S) = \frac{1}{N}\sum_{j=1}^N \|\hat{t}(x_j)-\hat{s}(x_j)\|_2^2
\end{align}
for our student training, where $\hat{t}(x_j)$ and $\hat{s}(x_j)$ are the teacher and the student logit vectors for input $x_j$.
Data augmentation plays a crucial role in our method. Our data augmentation scheme is as follows. To every image we apply a standard ``base geometric transform''. On top of this, we apply further $n_{ops}$ operations, chosen randomly, out of a pool of $n_{tot}$ operations. Please refer to Appendix for more details. Before feeding the images to the student and teacher, we normalize them by the mean $\mu_{synth}$ and standard deviation $\sigma_{synth}$ of the synthetic dataset.:
\begin{align*}
x_i \to \frac{x_i-\mu_{synth}}{\sigma_{synth}}
\end{align*}
Since ZOPGA-KD is a data-free KD method, we have not used mean and standard deviation of the original training set. We emphasize that during the student training the \textit{teacher is queried again} to get the logits for the normalized generated data. The student training protocol is schematically presented in Fig.~\ref{fig:student_training}.

\begin{figure*}[t]
    \centering
    \includegraphics[width=\linewidth]{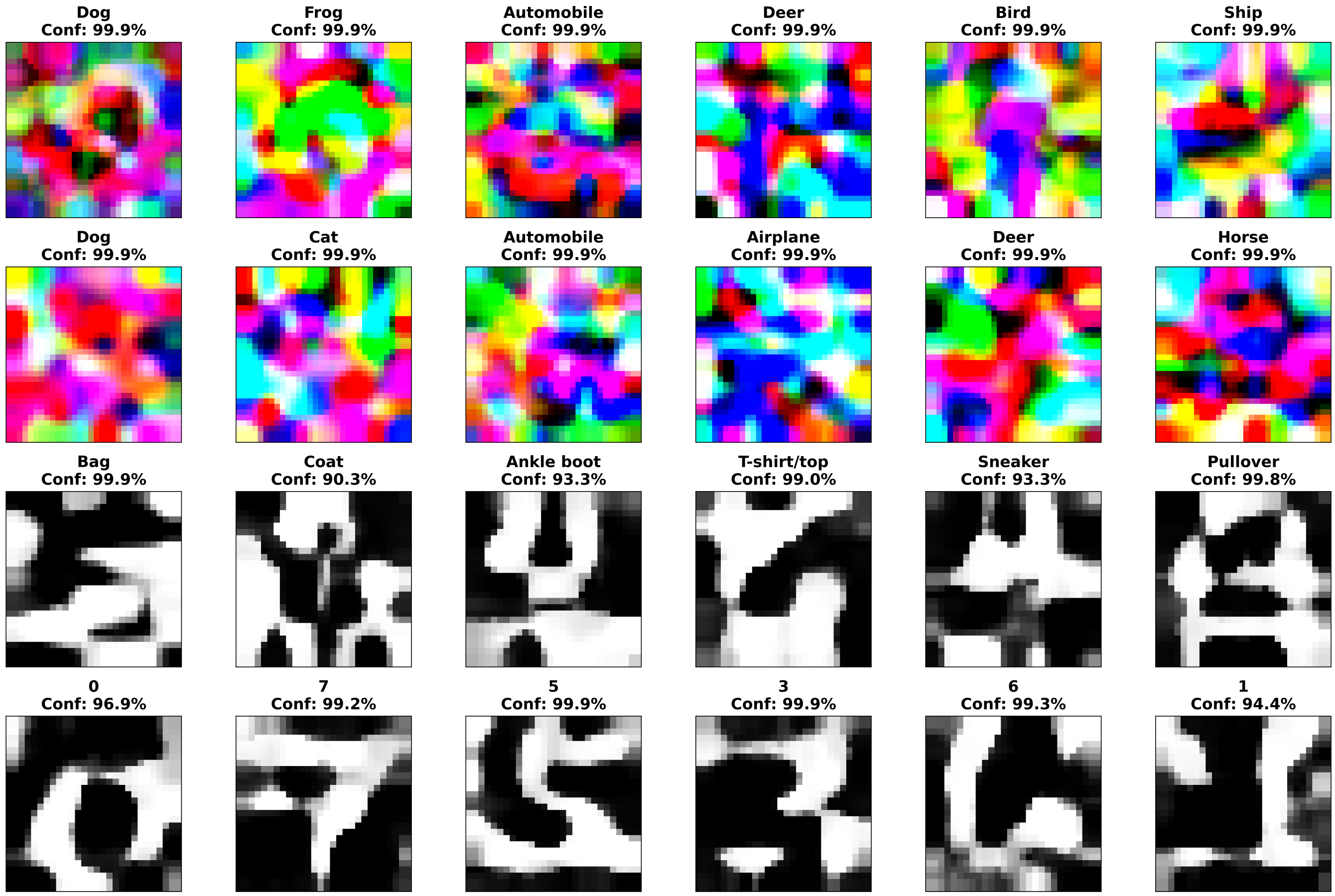} 
    \caption{Representative synthetic samples generated by ZOPGA-KD. From top to bottom, the rows correspond to the AlexNet/CIFAR-10, ResNet/CIFAR-10, LeNet/Fashion-MNIST, and LeNet/MNIST teacher-to-student configurations.}\label{fig:synthetic_samples}
\end{figure*}

\section{Experimental Results}\label{sec:experiments}

\subsection{Experimental Setup}

We evaluate the proposed black-box data-free knowledge distillation framework ZOPGA-KD on three representative teacher $\rightarrow$ student settings spanning shallow and deep convolutional neural networks:

\begin{itemize}
    \item \textbf{LeNet-5 $\rightarrow$ LeNet-5-Half} on MNIST \cite{deng2012mnist} and Fashion-MNIST \cite{xiao2017fashion} datasets,
    \item \textbf{AlexNet $\rightarrow$ AlexNet-Half} on CIFAR-10 \cite{krizhevsky2009learning},
    \item \textbf{ResNet-34 $\rightarrow$ ResNet-18} on CIFAR-10.
\end{itemize}

The complete experimental pipeline is identical across all datasets and architectures, differing only in the underlying network definitions and architecture-specific training hyperparameters, which is described in the Appendix. Each experiment consists of three steps: (i) supervised teacher training on the original dataset, (ii) black-box synthetic image generation using the proposed ZO-PGA algorithm, and (iii) data-free knowledge distillation by training the student exclusively on the generated synthetic dataset. After every epoch of the student training, we evaluate the student on the real test set and report the overall best accuracy. This is a standard practice in the literature \cite{zhang2022ideal, rosenthal2023disguide, nayak2019zero,wang2021data,wang2021zero}.


Teacher models are trained using standard cross-entropy loss until convergence. Student models in the proposed DFKD setting are trained with MSE loss function, exclusively using the synthesized dataset without access to any real training images. The teachers are trained with normalized inputs with dataset specific normalizations. During the data generation, however, we use unnormalized inputs. Finally, during the student training we normalize the inputs using the statistics of the synthetic data. We compare our results with the \citet{nayak2019zero,wang2021zero} for LeNet over MNIST, FMNIST and AlexNet over CIFAR-10, and \citet{wang2021data} for LeNet over MNIST and AlexNet, ResNet over CIFAR-10. We compare against these methods because, like ours, they are generator-free and synthesize the  dataset by directly optimizing samples in the input space, without training a generative network. We do not compare our results with generator based models such as IDEAL (\citet{zhang2022ideal}), MAZE (\citet{kariyappa2021maze}) and DFME (\citet{truong2021data}). More details about specific hyperparameters and training protocols can be found in Appendix.

\subsection{Synthetic Data Generation}

Synthetic images were generated using the algorithm described in Sec.~\ref{sec:method}. For all the experiments, random smooth noise signals are utilized with constant gradients along random directions as our initial noise. Unless otherwise specified, we use the upper bound on the confidence score $\tau=0.999$, iteration limit $t_{\max}=2000$, momentum $\beta=0.9$, pool size $B=256$, directions per step $q=8$, and smoothing parameter $\sigma=0.01$. We also apply a cosine learning-rate scheduler that interpolates between learning rates $1$ and $0.1$. The number of generated synthetic data samples for each of our experiments are presented in the Table \ref{tab:main-results}.

\subsection{Data Augmentation}
Our data augmentation scheme is described in Sec.~\ref{sec:method}. We first apply a base geometric transformation followed by taking $n_{ops} = 4$ additional operations chosen randomly from a pool of $n_{tot} = 17$ operations. For base geometric transformation we use reflected padding with padding=2 followed by a random horizontal flip except for the MNIST case for which the horizontal flip is disabled.

\subsection{Query Budget}
We query the teacher in two disjoint phases, reported separately in the lower block of Table ~\ref{tab:main-results}.\ \emph{Generation} queries are spent synthesizing the $N_{\mathrm{synth}}$ images. The count is data-dependent because abandonment, restarts, and adaptive resolution make the per-sample cost depend on how quickly a candidate reaches $\tau$.\ \emph{Augmentation} queries arise during online student distillation: each augmented image is a new input, so the teacher must recalculate its logits (see Fig.~\ref{fig:student_training}). This gives $200\times N_{\mathrm{synth}}$ queries over 200 epochs, where $N_{\mathrm{synth}}$ is the number of synthetic samples. The number of queries is sensitive to the parameter values reached by the teacher during training because the functions $T_i:\text{input space}\to[0,1]$, which PGA seeks to maximize, depend on the teacher parameters. The ZSDB3KD query costs in Table~\ref{tab:main-results} are lower bounds on the
total number of teacher queries. For each of the $N$ pseudo-samples, the cost is
the sum of (a) $I \times B$ queries for generation,where $I$ is the number of iteractions required for moving the sample away from the decision boundaries, and $B$ is the query budget for each iteration (b)
$(K-1) \times B$ queries for soft-label construction where $K$ is the number of classes. The total is therefore
$N \times (I + K - 1) \times B$. The paper uses $N=80000$ and $B=5000$. \emph{Queries used for gathering the initial samples are excluded}.

We evaluate our method using the classification accuracy of the distilled student on the real testsets of the respective datasets. For comparison, we also report the teacher accuracy, the accuracy of a student trained from scratch using cross-entropy loss, the accuracy obtained through classical knowledge distillation, and the total number of teacher queries used for synthetic-data generation and augmentation.

\subsection{Results}



Table~\ref{tab:main-results} reports student test accuracy across the four teacher $\rightarrow$ student configurations, together with the teacher-query budget
consumed by ZOPGA-KD\@. Despite operating in the black-box setting,
ZOPGA-KD recovers most of the performances corresponding to the teacher accuracy on LeNet/MNIST, LeNet/Fashion-MNIST, AlexNet/CIFAR10, and ResNet/CIFAR10, respectively.

Compared with ZSDB3KD, the only black-box, data-free baseline,
ZOPGA-KD achieves slightly improved accuracy on LeNet/MNIST ($97.40\%$ vs.\
$96.54\%$) and substantially higher accuracy on the other evaluated settings:
$80.06\%$ vs.\ $72.31\%$ on LeNet/Fashion-MNIST and $78.41\%$ vs.\ $59.46\%$ on
AlexNet/CIFAR-10, corresponding to improvements of $7.75\%$ and $18.95\%$, respectively. More notably, ZOPGA-KD outperforms even the white-box data-free methods like ZSKD and \citet{wang2021data}. On AlexNet/CIFAR10 it exceeds both ZSKD ($69.56\%$) by 8.85\% and \citet{wang2021data} ($73.91\%$) by 4.5\%. It further matches DB3KD-MBD on AlexNet/CIFAR-10 ($78.41\%$ vs.\ $78.30\%$) and exceeds it by $10.66\%$ on ResNet/CIFAR-10, even though the student network have access to real training data. We present a representative set of synthetic samples from the ZOPGA-KD generation process in Fig.~\ref{fig:synthetic_samples}. 


\section{Conclusion}

We propose a novel data-free knowledge distillation method (ZOPGA-KD) that utilizes Zeroth-Order PGA to generate synthetic data, which is then used to train the student network. The proposed method assumes that the teacher network parameters and original data on which the teacher is trained are not available due to privacy or any other reasons. ZOPGA-KD results are evaluated for \textbf{LeNet-5 $\rightarrow$ LeNet-5-Half} on MNIST and Fashion-MNIST, \textbf{AlexNet $\rightarrow$ AlexNet-Half} and \textbf{ResNet-34 $\rightarrow$ ResNet-18} on CIFAR-10. The proposed method outperforms existing generator-free black-box approaches, achieving performance comparable to—and occasionally exceeding—state-of-the-art white-box methods.

Two obvious future directions of our work would be to a). generalize our method to the hard label setting and b). extend it to more complex datasets such as CIFAR-100 and TinyImageNet.  Another important direction would be to further optimize the data generation algorithm to reduce the number of queries. In Table~\ref{tab:main-results} we observe that the number of queries required to get high confidence data samples is highly sensitive to the model architecture. Overall, these findings provide a strong basis for future investigations into this phenomenon.

\section*{Appendix}

\begin{figure}[h]
    \centering
    \includegraphics[width=0.2\textwidth]{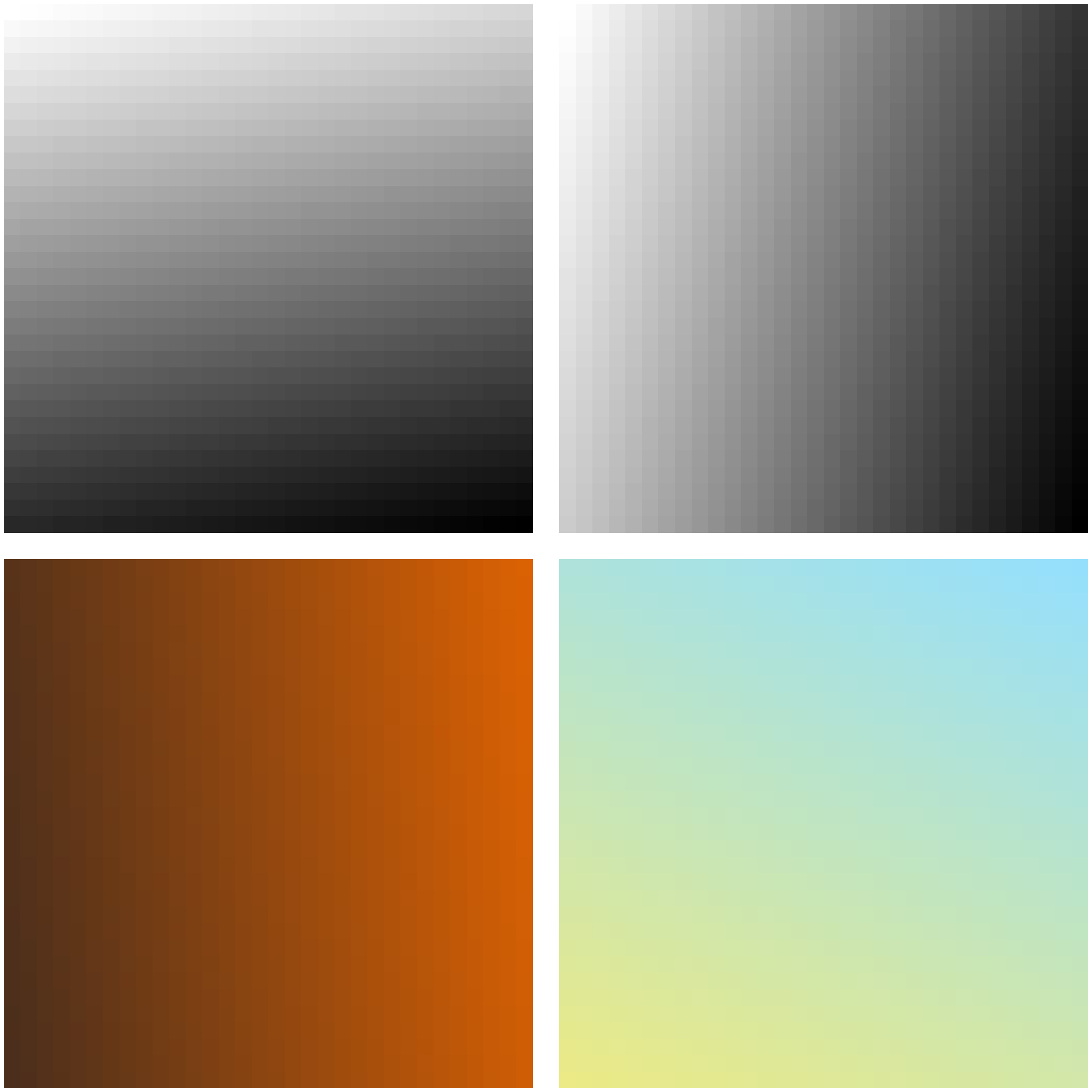}
    \caption{A few samples of initial smooth gradient noise images.}
    \label{fig:init_image}
    
\end{figure}

\section{Initial Noise Images}
\label{app:initialization}
ZOPGA works on initial noise images to evolve them into a desired synthetic dataset with high-confidence samples. Fig.~\ref{fig:init_image} shows a few such initial samples. These images are generated by random constant color gradients along random directions.

\section{Data Augmentation}
In this section, we describe the data augmentations used during the student training. As mentioned in the manuscript, we use two base geometric transformations followed by $n_{ops}=4$ operations chosen from a set of $n_{tot}=17$ options (see Ref.~\citep{cubuk2020randaugment}). The exact implementations of these transformations can be found in Ref.~\cite{augment}.

\subsection*{Base Geometric transformations}
We use the following base geometric transformations:
\begin{itemize}
\item Random crop with reflection padding
\item Random horizontal flip
\end{itemize}
For MNIST, we switch off the horizontal flip.
\subsection*{Additional Operations}
On top of the base geometric transformations, we apply four additional operations chosen from the set of 17 operations presented in Table~\ref{tab:transformations}. 


\begin{strip}
\centering
\small
\captionof{table}{Summary of Data Augmentations and Image Transformations.}
\label{tab:transformations}
\begin{tabular}{ll}
\toprule
\textbf{Category} & \textbf{Transformations} \\
\midrule
\textbf{Geometric} & Rotation, Affine transformation, Change of perspective, Zoom-crop \\
\addlinespace
\textbf{Photometric \& Color} & Color jitter, Grayscale conversion, Sharpness adjustment, \\
& Contrast adjustment, Equalize, Posterize, Solarize, Pixel inversion \\
\addlinespace
\textbf{Blur \& Noise} & Gaussian blur, Gaussian noise, Salt-and-pepper noise \\
\addlinespace
\textbf{Erasing \& Occlusion} & Random pixel erasing, Replacing random patches with random colors \\
\bottomrule
\end{tabular}
\end{strip}

\section{Hyperparameter Details}
We train all of our student models for 200 epochs. The confidence threshold $\tau$ and the maximum number of steps $T_{max}$ are chosen to be 0.999 and 2000 steps, respectively. In Table~\ref{tab:hyperparameters}, we present the batch sizes for the student training with synthetic data. Additional hyperparameter details are provided in Table~\ref{tab:pipeline_hyperparams}.

\begin{table}[t]
\centering
\small
\caption{Hyperparameter configurations for our experiments.}
\label{tab:hyperparameters}
\begin{tabular}{lcc}
\toprule
\textbf{System} & \textbf{Samples ($N_{\mathrm{synth}}$)} & \textbf{Batch Size} \\
\midrule
\textbf{LeNet} (MNIST)     & 24K & 16 \\
\textbf{LeNet} (FMNIST)    & 48K & 16 \\
\textbf{AlexNet} (CIFAR10) & 40K / 200K$^\dagger$ & 128 \\
\textbf{ResNet} (CIFAR10)  & 40K / 200K$^\dagger$ & 512 \\
\bottomrule
\multicolumn{3}{l}{\footnotesize $^\dagger$Samples used in additional student experiments.}
\end{tabular}
\end{table}

\begin{table*}[t]
\centering
\small
\caption{Additional hyperparameter settings for the ZOPGA pipeline and student training.}
\label{tab:pipeline_hyperparams}
\begin{tabular}{p{0.23\textwidth}p{0.42\textwidth}p{0.25\textwidth}}
\toprule
\textbf{Phase} & \textbf{Hyperparameter} & \textbf{Value} \\
\midrule
\multirow{3}{=}{\textbf{ZO-PGA Generation}}
  & Step Size ($\alpha$) / Smoothing Parameter ($\sigma$) & $0.5$ / $0.01$ \\
  & Queries per Step ($q$) / Confidence Threshold ($\tau$) & $8$ / $0.999$ \\
  & EMA Momentum & $0.9$ \\
\midrule
\multirow{2}{=}{\textbf{Student DFKD}}
  & Initial LR / Weight Decay & $0.001$ / $5 \times 10^{-4}$ \\
  & LR Scheduler ($\text{Step}$/LR-decay) & StepLR ($25$ epochs / $0.5$) \\
\midrule
\multirow{3}{=}{\textbf{Student CE and KD Baselines}}
  & Initial LR/Weight Decay & $0.001$/$5 \times 10^{-4}$\\
  & LR Scheduler (step/LR-decay) &  StepLR ($25$ epochs / $0.5$) \\
  & KD Loss Temperature & $T = 20.0$ \\
\bottomrule
\end{tabular}
\end{table*}

\begin{table}[t]
\centering
\footnotesize
\caption{Queries and accuracy metrics across additional runs.}
\label{tab:queries_and_accuracies}
\resizebox{\columnwidth}{!}{%
\begin{tabular}{lccc}
\toprule
\textbf{Metric / Setting} & \textbf{AlexNet (200K)} & \textbf{ResNet (200K)} & \textbf{ResNet ($40K,\tau=0.6$)} \\
\midrule
Batch Size ($\text{BS}$) & 128 & 512 & 512 \\
Queries used & 921.34M & 161.41M & \textbf{15.68M} \\
\midrule
Teacher-CE (\%) & 84.93 & 95.13 & 95.31 \\
Student-CE (\%) & 84.76 & 94.53 & 93.93 \\
Student-KD (\%) & 82.81 & 94.57 & 93.90 \\
ZOPGA-KD (\%) & 79.81 & \textbf{91.32} & 82.45 \\
\bottomrule
\end{tabular}%
}
\end{table}

\section{Additional Runs}
Table~\ref{tab:queries_and_accuracies} summarizes our additional experiments. We evaluate ZOPGA-KD on AlexNet-CIFAR10 and ResNet-CIFAR10 with a larger synthetic dataset of 200K samples. We also evaluate ResNet with 40K samples and a lower confidence threshold ($\tau=0.6$). With 200K synthetic samples, ResNet reaches 91.32\% accuracy, a $3.34$-point improvement over the 40K-sample setting. At $\tau=0.6$, accuracy reaches 82.45\% while the query budget drops by 56.61\% and accuracy falls by 5.52 percentage points.

{\hbadness=2000\bibliography{aaai2027}}
\end{document}